\documentclass{article}
\usepackage{spconf,amsmath,graphicx,cite,tcolorbox,amsfonts,booktabs,multirow,subcaption,adjustbox,array,makecell,enumitem,adjustbox,afterpage}
\usepackage{url}
\tcbuselibrary{breakable}

\title{SemanticShield: LLM-Powered Audits Expose Shilling Attacks in Recommender Systems}
\name{Kaihong Li$^{1}$, Huichi Zhou$^{2}$, Bin Ma$^{3}$, Fangjun Huang$^{1*}$}
\address{
$^{1}$ Sun Yat-sen University,
$^{2}$ University College London,
$^{3}$ Qilu University of Technology 
}
\begin{document}
\ninept
\maketitle
\begin{abstract}
Recommender systems (RS) are widely used in e-commerce for personalized suggestions, yet their openness makes them susceptible to shilling attacks, where adversaries inject fake behaviors to manipulate recommendations. Most existing defenses emphasize user-side behaviors while overlooking item-side features such as titles and descriptions that can expose malicious intent. To address this gap, we propose a two-stage detection framework that integrates item-side semantics via large language models (LLMs). The first stage pre-screens suspicious users using low-cost behavioral criteria, and the second stage employs LLM-based auditing to evaluate semantic consistency. Furthermore, we enhance the auditing model through reinforcement fine-tuning on a lightweight LLM with carefully designed reward functions, yielding a specialized detector called SemanticShield. Experiments on six representative attack strategies demonstrate the effectiveness of SemanticShield against shilling attacks, and further evaluation on previously unseen attack methods shows its strong generalization capability. Code is available at~\url{https://github.com/FrankenstLee/SemanticShield}.    
\end{abstract}

%

%
\section{Introduction}

\label{sec:intro}

Recommender systems have become indispensable in mitigating information overload, enabling users to access relevant content through personalized suggestions \cite{wu2021survey,koren2021advances}. Collaborative filtering (CF) has long been the dominant paradigm \cite{hu2008collaborative,he2020lightgcn}, but its reliance on user data also exposes it to adversarial manipulations, particularly shilling attacks \cite{gunes2014shilling}, where fake user profiles are injected to promote target items and undermine system reliability. To address this issue, numerous detection methods have been proposed. For example, Mehta and Nejdl~\cite{mehta2009unsupervised} applied unsupervised dimensionality reduction techniques to separate anomalous user profiles from genuine ones in the latent space, while DeR-TIA~\cite{zhou2014detection} employed rating deviation metrics together with target item analysis to identify coordinated group attacks.
Building on user–item interactions, CoDetector~\cite{dou2017collaborative} constructed a co-visitation graph to capture suspicious behaviors and item-level collusion patterns. More recently, DGA-MFCA~\cite{xu2024detecting} leveraged multi-dimensional user representations and synchronized rating behaviors to detect subtle collusion. Despite their effectiveness against classical attacks, these approaches still face notable limitations: (1) reliance on heuristic priors, limiting generalization to novel strategies such as reinforcement learning–based attacks
\cite{song2020poisonrec}; (2) exclusive reliance on user-side behaviors, overlooking item-side semantics that could reveal incoherent user activities\cite{sun2019research}; and (3) high false-positive rates, which degrade user experience.

Recent advances in large language models (LLMs) provide new opportunities for attack detection \cite{zhao2023survey}. LLMs excel at semantic reasoning and contextual understanding, and have shown effectiveness in recommender-related tasks \cite{wu2024survey,zhao2024recommender}. Building on these strengths, we propose a two-stage ``behavioral pre-screening and semantic auditing'' framework. In the first stage, a PCA-based \cite{abdi2010principal} similarity filter is applied to flag users with unusually high similarity to others, followed by an unpopular-item ratio filter used to identify users who overly focus on low-popularity items. In the second stage, LLMs audit user interaction histories to further screen suspicious users. Furthermore, reinforcement fine-tuning (RFT) is adopted to enhance detection performance. Unlike supervised fine-tuning, which only memorizes the data patterns~\cite{chu2025sft}, RFT can achieve the generalization based on rule-based reward signal in different downstream tasks, such as Group Relative Policy Optimization (GRPO)~\cite{shao2024deepseekmath}. Building on this, we fine-tune Qwen2.5-1.5B-Instruct\cite{qwen2.5} with carefully designed reward functions for malicious user detection, thereby deriving SemanticShield. SemanticShield surpasses Llama-3-70B-Instruct\cite{llama3modelcard} in auditing accuracy, yielding a robust and specialized detector against diverse shilling attacks.


Our contributions are three-fold: 
\textbf{(1)} We propose a novel LLM-based detection framework that incorporates item-side semantics into shilling attack identification, addressing a critical gap in prior work. \textbf{(2)} We conduct comprehensive experiments on real-world datasets across multiple representative attack strategies, demonstrating the strong generalization ability of LLM-based detection. \textbf{(3)} We fine-tune Qwen2.5-1.5B-Instruct using GRPO reinforcement learning, yielding an expert-level auditing model that consistently surpasses state-of-the-art baselines in detection effectiveness.


\section{Approach}
\label{sec:majhead}

\subsection{Definition of Shilling Attack}
\label{sec:shillingattack}

Let $\mathcal{U}$ and $\mathcal{I}$ denote the sets of users and items, respectively. The user–item interactions can be represented as a matrix $R \in \mathbb{R}^{|\mathcal{U}|\times|\mathcal{I}|}$, where $R_{ui} > 0$ indicates that user $u$ has interacted with item $i$. A recommender system then trains a model $f_{\theta}$ on $R$ to learn user and item embeddings  for predicting user–item interactions, formally defined as:
\begin{equation}
\theta^{*} = \arg\min_{\theta} \mathcal{L}_{\text{rec}}\big(f_{\theta}(R)\big),
\label{eq:rec_obj}
\end{equation}
where $\theta^{*}$ is the learned parameter set of the recommender, and $\mathcal{L}_{\text{rec}}$ denotes the recommendation objective function, e.g., the Bayesian Personalized Ranking (BPR) loss \cite{rendle2012bpr}.  

In a shilling attack, an adversary manipulates a subset of users $\mathcal{U}_M$ by generating synthetic interactions $R_M$ and injecting them into the training process. The recommendation model is subsequently optimized on the contaminated data ${R} \cup {R}_M$, which can be formulated as a bi-level optimization problem:  
\begin{equation}
\begin{aligned}
{R}_M &= \arg\max_{{R}_M} \mathcal{L}_{\text{attack}}\!\left({R} \cup {R}_M, \theta^{*}\right), \\
\text{s.t.} \quad
\theta^{*} &= \arg\min_{\theta} \mathcal{L}_{\text{rec}}\!\left(f_{\theta}({R} \cup {R}_M)\right).
\end{aligned}
\label{eq:attack_obj}
\end{equation}
where $\mathcal{L}_{\text{attack}}$ represents the attacker’s objective function, which measures the success of poisoning in terms of promoting the ranking positions of a target item set $\mathcal{I}_t \subseteq \mathcal{I}$.



\subsection{Stage I: Behavior-Only Pre-screening}
\label{ssec:subhead1}

To alleviate the burden on the downstream LLM auditor, a candidate set of potentially suspicious users is first identified based solely on rating behaviors. 
To this end, the following filtering methods are applied to the full interaction data \( {R} \cup {R}_M \) in this stage.

\textbf{PCA Similarity Filter.}  Each user’s interaction vector is projected via PCA, and similarity between all user pairs is computed. Users whose similarity to any other user exceeds a threshold \( \delta \) are flagged:
\begin{equation}
\mathcal{S}_{\text{PCA}} = \left\{\, u \in \mathcal{U} \;\middle|\; \exists v \in \mathcal{U},\, v \ne u,\ s(u,v) > \delta \,\right\},
\label{eq:spca}
\end{equation}
where \( s(u,v) \) denotes the cosine similarity between users \( u \) and \( v \) in the PCA-projected space.

\textbf{Unpopular-Item Ratio Filter.} Define item popularity as \( \mathrm{pop}(i) = \sum_{u \in \mathcal{U}} \mathbb{I}[R_{ui} > 0] \), and let \( \mathcal{I}_{\text{unpop}} \) denote the set of items in the lowest popularity percentile. For each user \( u \), let \( \mathcal{I}(u) = \{\, i \in \mathcal{I} \mid R_{ui} > 0 \,\} \) be the set of items interacted with by \( u \). Users whose unpopular-item ratio exceeds threshold \( \tau \) are flagged:
\begin{equation}
\mathcal{S}_{\text{UNPOP}} = \left\{\, u \in \mathcal{U} \;\middle|\; \frac{|\mathcal{I}(u) \cap \mathcal{I}_{\text{unpop}}|}{|\mathcal{I}(u)|} \ge \tau \,\right\}.
\label{eq:sunpop}
\end{equation}

The final candidate set is formed as \( \mathcal{S} = \mathcal{S}_{\text{PCA}} \cup \mathcal{S}_{\text{UNPOP}} \), and these users are passed to the LLM auditor, while those in \( \mathcal{U} \setminus \mathcal{S} \) are considered genuine.

\subsection{Stage II: Open-Source LLM Auditing}
\label{ssec:subhead2}

Given the candidate set $\mathcal{S}$ from Stage~I, we employ an LLM (Llama-3-70B-Instruct) to further audit user behavior.  
To standardize the interaction with the LLM, the following prompt template is designed:

\begin{tcolorbox}[
  colback=gray!5,
  colframe=gray!40!black,
  title=Prompt Template for LLM,
  boxrule=0.5pt,
  breakable,
  left=0pt,
  right=0pt,
  top=2pt,
  bottom=2pt
]
\textbf{System prompt:} \\
\textit{Role:} You are a behavioral auditor within an adversarial recommender system. \\
\textit{Guidelines:} \textless\textnormal{Prior knowledge}\textgreater \\
Report the following: \\
1. Detailed reasoning process. \\
2. \textless\textnormal{Judgment format}\textgreater \\
Please respond in exactly the following format: \\
\textless\textnormal{Response template}\textgreater 
\\[3pt]
\textbf{User content:} \\
Item$_1$ (features); Item$_2$ (features); $\ldots$; Item$_N$ (features).
\end{tcolorbox}

In the above prompt, \textless Prior knowledge\textgreater\ offers context about genuine user behaviors within the specific recommender system.  
\textless Judgment format\textgreater\ specifies the judgment format to be provided, and \textless Response template\textgreater\ defines a structured answer format to ensure consistency and facilitate reliable parsing.

In Stage~II, the \textless Judgment format\textgreater\ requires the LLM to return a confidence score from $1$ to $5$, indicating the likelihood that a given user is genuine.  
This design leverages the larger parameter scale of Llama-3-70B-Instruct to handle more complex reasoning tasks, thereby reducing the risk of misclassification.

Based on this prompt, for each user $u \in \mathcal{S}$, an interaction sequence $s_u$ is constructed by extracting all items $i$ with $R_{ui} > 0$ from the user’s interaction history $R_u$. For each such item, the corresponding item-side features (e.g., titles and descriptions) are also retrieved and concatenated with the interaction sequence. The LLM then analyzes $s_u$ under the guidance of \textless Prior knowledge\textgreater\ and returns a confidence score $r_u$ ranging from 1 to 5, where higher values indicate stronger alignment with benign behavior. Binary labels are assigned by thresholding: users with $r_u \geq 3$ are regarded as genuine, while those with $r_u < 3$ are regarded as fake. The final set of malicious users is thus defined as $\mathcal{F} = \{ u \in \mathcal{S} \mid r_u < 3 \}$.

\newcolumntype{C}[1]{>{\centering\arraybackslash}p{#1}}

\begin{table*}[t!]
\centering
\normalsize
\setlength{\tabcolsep}{1.5pt}
\renewcommand{\arraystretch}{1.1} 
\caption{\centering \textbf{
Detection performance (DR / FAR) of all methods against six representative shilling attacks on ML-1M, MIND, and Clothing datasets. Average results are included to assess overall detection robustness, and the best results are in bold.}
}
\begin{adjustbox}{max width=\textwidth}
\begin{tabular}{C{1.4cm} C{3.0cm} *{14}{C{1.2cm}}}
\toprule
\multirow{2}{*}{\textbf{Dataset}} & \multirow{2}{*}{\textbf{Method}}
& \multicolumn{2}{c}{\textbf{Random(\%)}}
& \multicolumn{2}{c}{\textbf{Bandwagon(\%)}}
& \multicolumn{2}{c}{\textbf{GTA(\%)}}
& \multicolumn{2}{c}{\textbf{PoisonRec(\%)}}
& \multicolumn{2}{c}{\textbf{PipAttack(\%)}}
& \multicolumn{2}{c}{\textbf{CLeaR(\%)}}
& \multicolumn{2}{c}{\textbf{Avg.}(\%)}\\
\cmidrule(lr){3-4}\cmidrule(lr){5-6}\cmidrule(lr){7-8}\cmidrule(lr){9-10}\cmidrule(lr){11-12}\cmidrule(lr){13-14}\cmidrule(lr){15-16}
& & \textbf{DR↑} & \textbf{FAR↓} & \textbf{DR↑} & \textbf{FAR↓} & \textbf{DR↑} & \textbf{FAR↓} & \textbf{DR↑} & \textbf{FAR↓} & \textbf{DR↑} & \textbf{FAR↓} & \textbf{DR↑} & \textbf{FAR↓} & \textbf{DR↑} & \textbf{FAR↓}\\
\midrule
\multirow{6}{*}{\textbf{ML-1M}}
& \textbf{PCA-VarSelect}          & 53.33 & 16.04 & 75.00 & 16.37 & 0.00 & 17.17 & 0.00 & 17.02 & 0.00 & 17.07 & 0.00 & 17.09 & 21.39 & 16.79\\
& \textbf{CBS}                & 93.33 & 1.59 & 95.00 & 1.59 & 95.00 & 1.59 & 95.00 & 1.59 & 96.67 & 1.59 & 95.00 & 1.59 & 95.00 & 1.59\\   
& \textbf{GAGE}                & 0.00 & 27.63 & 100.00 & 23.21 & 95.00 & 24.85 & 46.67 & 19.22 & 0.00 & 27.15 & 0.00 & 27.17 & 40.28 & 24.87\\
& \textbf{DGA-MFCA}               & 100.00 & 11.95 & 28.33 & 11.89 & 100.00 & 11.69 & 100.00 & \textbf{0.00} & 10.00 & 12.40 & 15.00 & 12.40 & 58.89 & 10.06\\
& \textbf{Llama-3-70B-Instruct}   & 89.66 & 0.72 & 93.33 & 0.71 & 96.67 & 0.70 & 91.53 & 0.71 & 93.33 & 0.75 & 91.67 & 0.73 & 92.70 & 0.72\\
    & \textbf{SemanticShield}         & \textbf{100.00} & \textbf{0.10} & \textbf{100.00} & \textbf{0.10} & \textbf{100.00} & \textbf{0.07} & \textbf{100.00} & 0.07 & \textbf{100.00} & \textbf{0.05} & \textbf{100.00} & \textbf{0.03} &\textbf{100.00} & \textbf{0.07}\\
\midrule
\multirow{6}{*}{\textbf{MIND}}
& \textbf{PCA-VarSelect}          & 96.67 & 44.01 & 0.00 & 46.18 & 0.00 & 42.35 & 100.00 & 43.59 & 3.33 & 44.52 & 5.00 & 44.67 & 34.17 & 44.22\\
& \textbf{CBS}                    & 93.33 & 1.59 & 81.67 & 1.74 & 96.67 & 1.59 & 86.67 & 1.66 & 80.00 & 1.71 & 80.00 & 1.74 & 86.39 & 1.67\\
& \textbf{GAGE}                & 91.67 & 0.33 & 96.67 & 0.40 & 90.00 & 0.33 & 83.33 & 29.69 & 68.33 & 0.35 & 70.00 & 0.40 & 83.33 & 5.25\\
& \textbf{DGA-MFCA}               & 0.00 & 32.96 & 100.00 & 9.16 & 100.00 & \textbf{0.00} & 100.00 & 27.10 & 95.00 & 14.97 & 93.33 & 18.33 & 81.39 & 17.09\\
& \textbf{Llama-3-70B-Instruct}  & 85.00 & 0.73 & 96.67 & 0.88 & 98.33 & 0.71 & 91.67 & 1.12 & 68.97 & 1.22 & 75.86 & 1.18 & 86.08 & 0.97\\
& \textbf{SemanticShield}          & \textbf{100.00} & \textbf{0.22} & \textbf{100.00}& \textbf{0.22} & \textbf{100.00} & 0.25 & \textbf{100.00} &\textbf{0.23} & \textbf{98.33} & \textbf{0.26} & \textbf{98.33} & \textbf{0.23} & \textbf{99.44} & \textbf{0.24}\\
\midrule
\multirow{6}{*}{\textbf{Clothing}}
& \textbf{PCA-VarSelect}          & 78.33 & 31.01 & 0.00 & 31.79 & 0.00 & 31.56 & 80.00 & 31.09 & 0.00 & 32.35 & 0.00 & 32.30 & 26.39 & 31.68\\
& \textbf{CBS}                   & 95.00 & 1.59 & 80.00 & 1.74 & 93.33 & 1.59 & 26.67 & 2.32 & 88.33 & 1.66 & 78.33 & 1.74 & 76.94 & 1.77\\ 
& \textbf{GAGE}                & 85.00 & 17.42 & 88.33 & 1.04 & 96.67 & 0.89 & 86.67 & 17.12 & 90.00 & 1.24 & 86.67 & 1.04 & 88.89 & 6.46\\
& \textbf{DGA-MFCA}               & 0.00 & 7.68 & 96.67 & 9.29 & 100.00 & 3.96 & 0.00 & 7.68 & 98.33 & 6.92 & 95.00 & 10.35 & 65.00 & 7.65\\
& \textbf{Llama-3-70B-Instruct}   & 98.31 & 2.23 & 78.33 & 2.34 & 100.00 & 2.68 & 95.00 & 2.41 & 100.00 & 2.20 & 77.97 & 1.87 & 91.60 & 2.29\\
& \textbf{SemanticShield}           & \textbf{100.00} & \textbf{0.53} & \textbf{98.33} & \textbf{0.48} & \textbf{100.00} & \textbf{0.51} & \textbf{100.00} & \textbf{0.53} & \textbf{100.00} & \textbf{0.48} & \textbf{100.00} & \textbf{0.53} & \textbf{99.72} & \textbf{0.51}\\
\bottomrule
\end{tabular}
\end{adjustbox}

\label{tab:main_results}
\end{table*}

\subsection{SemanticShield: Fine-Tuning via GRPO}
\label{ssec:grpo}

To improve the classification accuracy in Stage~II, the Qwen2.5-1.5B-Instruct model is fine-tuned using GRPO, resulting in our specialized detector SemanticShield. The prompt schema used in this stage is consistent with the template introduced in \S~\ref{ssec:subhead2}, where the \textless Judgment format\textgreater{} requires the model to output a discrete label of either \textit{Real} or \textit{Fake}.  
This design simplifies the reasoning task for the smaller model and reduces the risk of overfitting during RFT.  
Specifically, the \textless Response template\textgreater{} is structured as follows:

\begin{quote}
\texttt{<think>} \emph{reasoning text} \texttt{</think>} \\
\texttt{<answer>} \emph{Real or Fake} \texttt{</answer>}
\end{quote}

To guide training, we design a composite reward signal that encourages outputs to be structurally valid, logically coherent, and behaviorally accurate. 
The reward consists of four components: 

A structure reward \( r_{\text{format}} \), which is given when the model's output strictly conforms to the \textless Response template\textgreater{} format. 

A clarity reward \( r_{\text{clarity}} \), which encourages interpretable reasoning by applying a regex-based matching of the \emph{reasoning text} extracted from the model's output. 
The model is rewarded if the reasoning follows an enumerated step format (e.g., 
1.~\emph{First step}; 2.~\emph{Second step}; 3.~\emph{Third step}; $\ldots$).

A consistency reward \( r_{\text{consist}} \), which penalizes cases where the reasoning content contradicts the final decision. 
For example, if the model's answer is \emph{Real} while the extracted \emph{reasoning text} explicitly states that the sample is \emph{Fake}, a penalty is applied.

 A task reward $r_{\text{task}}$, which serves as the primary supervision signal. It is obtained by comparing the predicted label $y$, which is extracted from the answer field of the model's output $o$, against the ground truth $y^\star$. 
The reward is formally defined as:
\begin{equation}
r_{\text{task}}(o, y^\star) =
\begin{cases}
+R_1, & \text{if } y = y^\star \\
-R_1, & \text{if } y \ne y^\star \text{ and } y^\star = \text{\emph{Real}} \\
-R_2, & \text{if } y \ne y^\star \text{ and } y^\star = \text{\emph{Fake}}
\end{cases}
\label{eq:task}
\end{equation}
with \( R_2 > R_1 > 0 \), assigning stronger penalties to false negatives. This design underscores the critical importance of accurately identifying malicious users.

Based on the prompt template and interaction history \( s_u \) from each user \( u \), a task-specific query \( q_u \) is constructed. For each query, the LLM generates a group of candidate outputs \( \{o_1, \dots, o_G\} \), with \( G \) denoting the maximum number of candidates, each representing a possible response under the current model. A scalar reward \( r_i \) is computed for each output \( o_i \), aggregated from the four reward components introduced above, as follows:
\begin{equation}
r_i = r_{\text{format}}(o_i) + r_{\text{clarity}}(o_i) + r_{\text{consist}}(o_i) + r_{\text{task}}(o_i, y^\star)
\label{eq:composite_reward}
\end{equation}

After obtaining the rewards, GRPO normalizes them within each sampled group and computes relative advantages using the group average as the baseline. The model is then updated to maximize the likelihood of outputs with higher relative rewards under KL regularization, thereby aligning generation with the reward signals.

\section{Results and Discussion}

\subsection{Experimental Datasets and Attack Settings}
In order to demonstrate the effectiveness of the proposed approach, experiments are conducted on the following datasets:

\noindent\textbf{ML-1M \cite{harper2015movielens}} The dataset contains 1{,}000{,}209 ratings from 6{,}040 users on approximately 3{,}900 movies. It includes movie information such as title, genre, and release year.

\noindent\textbf{MIND \cite{wu2020mind}} The dataset is the MIND-small subset with 50{,}000 users and approximately 39{,}860 news articles. Each article provides textual fields such as title, abstract and other related information.

\noindent\textbf{Clothing \cite{hou2024bridging}} The dataset is selected from the \textit{Clothing, Shoes and Jewelry} category of Amazon Reviews’23, which contains about 2.5 million users and 715 thousand products. The 10,560 most active users are retained to reduce computational costs. Each products provides metadata such as description, brand, and price.

In the attack scenario, LightGCN~\cite{he2020lightgcn} is adopted as the victim recommender model $f_\theta$, as defined in \S~\ref{sec:shillingattack}. For each dataset, five items are randomly selected to form the target item set $\mathcal{I}_t \subseteq \mathcal{I}$, and the number of malicious users $\mathcal{U}_M$ is defined as 1\% of the total number of genuine users. Each malicious user is assigned a fixed quota of synthetic interactions, determined by the average number of items interacted with by genuine users.

Based on this setup, six representative shilling attack methods are employed to generate fake users. The earlier methods such as RandomAttack~\cite{lam2004shilling} and BandwagonAttack~\cite{gunes2014shilling} are primarily heuristic-driven. In contrast, more recent approaches including PoisonRec~\cite{song2020poisonrec}, PipAttack~\cite{zhang2022pipattack}, GTA~\cite{wang2023revisiting} and CLeaR~\cite{wang2024unveiling} leverage deep learning techniques to construct more sophisticated and effective attacks.

\newcolumntype{C}[1]{>{\centering\arraybackslash}p{#1}}

\begin{table}[t!] 
    \centering 
    \scriptsize 
    \setlength{\tabcolsep}{3pt} 
    \renewcommand{\arraystretch}{1.3} 
    \caption{\centering \textbf{Generalization performance of SemanticShield on unseen attacks across ML-1M, MIND, and Clothing datasets.}} 
    \begin{tabular}{C{2.0cm} C{1.3cm} C{1.3cm} C{1.3cm} C{1.3cm}} 
        \toprule 
        \multirow{2}{*}{\textbf{Dataset}}
            & \multicolumn{2}{c}{\textbf{GOAT(\%)}} 
            & \multicolumn{2}{c}{\textbf{FedRecAttack(\%)}} \\ 
        \cmidrule(lr){2-3}\cmidrule(lr){4-5} 
            & \textbf{DR↑} & \textbf{FAR↓} & \textbf{DR↑} & \textbf{FAR↓} \\ 
        \midrule 
        \textbf{ML-1M}    & 100.00 & 0.07 & 100.00 & 0.10 \\ 
        \textbf{MIND}     & 100.00 & 0.25 & 98.33 & 0.23 \\ 
        \textbf{Clothing} & 96.67 & 0.51 & 98.33 & 0.51 \\ 
        \bottomrule 
    \end{tabular} 
    
    \label{tab:generalization} 
\end{table}

\subsection{Baseline Models and Metrics} 
The proposed approach is compared with the following baselines:

\noindent\textbf{PCA-VarSelect \cite{mehta2009unsupervised}} A method that detects anomalous users using dimensionality reduction techniques.

\noindent\textbf{CBS \cite{zhang2015catch}} A method that leverages the propagation of anomalous behavior to detect shilling attacks.

\noindent\textbf{GAGE \cite{zhang2020graph}} A method that applies graph embeddings and clustering techniques to detect group shilling attacks.

\noindent\textbf{DGA-MFCA \cite{xu2024detecting}} A method that models multi-dimensional user behaviors and captures collusive patterns via dual graph aggregation.

Detection performance is evaluated using Detection Rate (DR) and False Alarm Rate (FAR). DR measures the proportion of fake users correctly identified, while FAR measures the proportion of genuine users mistakenly classified as fake. In addition, recommendation quality is assessed with Hit Ratio (HR@\(N\)) and Normalized Discounted Cumulative Gain (NDCG@\(N\)), where \(N\) denotes the length of the top-\(N\) recommendation list. HR@\(N\) indicates whether the item a user actually interacted with appears in the top-\(N\) list, and NDCG@\(N\) further accounts for its position, assigning higher scores when the item ranks closer to the top.

    

\newcolumntype{C}[1]{>{\centering\arraybackslash}p{#1}}

\begin{table*}[t!]
\centering
\normalsize
\setlength{\tabcolsep}{2pt}
\renewcommand{\arraystretch}{1.25}
\caption{\centering \textbf{
Average recommendation consistency (RC$_{\text{HR}}$, RC$_{\text{NDCG}}$) before and after defenses across three datasets. 
Each cell shows the consistency percentage relative to clean training, with original HR@50 and NDCG@50 in parentheses.
}}
\begin{adjustbox}{max width=\textwidth}
\begin{tabular}{C{3.5cm} *{6}{C{2.8cm}}}
\toprule
\multirow{2}{*}{\textbf{Setting}} 
& \multicolumn{2}{c}{\textbf{ML-1M(\%)}} 
& \multicolumn{2}{c}{\textbf{MIND(\%)}} 
& \multicolumn{2}{c}{\textbf{Clothing(\%)}} \\
\cmidrule(lr){2-3}\cmidrule(lr){4-5}\cmidrule(lr){6-7}
& \textbf{RC$_{\text{HR}}$(HR@50)} & \textbf{RC$_{\text{NDCG}}$(NDCG@50)} 
& \textbf{RC$_{\text{HR}}$(HR@50)} & \textbf{RC$_{\text{NDCG}}$(NDCG@50)} 
& \textbf{RC$_{\text{HR}}$(HR@50)} & \textbf{RC$_{\text{NDCG}}$(NDCG@50)} \\
\midrule
\textbf{Clean}           & - (34.654) & - (25.358) & - (11.542) & - (6.408) & - (2.573) & - (1.481) \\
\textbf{Attack}          & 99.09\% (34.337) & 98.95\% (25.093) & 99.74\% (11.513) & 99.41\% (6.370) & 95.93\% (2.468) & 97.91\% (1.450) \\
\textbf{Llama-3-70B-Instruct} & 99.28\% (34.404) & 99.56\% (25.247) & 99.90\% (11.530) & 99.84\% (6.398) & 97.75\% (2.515) &  97.19\% (1.438)\\
\textbf{SemanticShield}  & 99.55\% (34.499) & 99.82\% (25.313) & 99.66\% (11.503) & 99.92\% (6.403) & 99.10\% (2.550) & 99.19\% (1.469) \\
\bottomrule
\end{tabular}
\end{adjustbox}

\label{tab:robustness}
\end{table*}

\subsection{Training Data Preparation}
Each clean dataset is first partitioned into training and test sets. For the training set, each of the six attack methods is executed three times, with target items drawn from the unpopular, popular, and random popularity categories, respectively, yielding three malicious user groups per method. Across all six methods and three datasets,  a total of 54 groups collectively constitute the complete set of malicious users for training, alongside an equal number of genuine users randomly sampled from each training set. For each user, we construct their interaction history, assign the appropriate label and use the resulting data to fine-tune the LLM as described in \S~\ref{ssec:grpo}.

\begin{figure}[t]
    \centering
    \adjustbox{center}{%
        \includegraphics[width=1.05\linewidth]{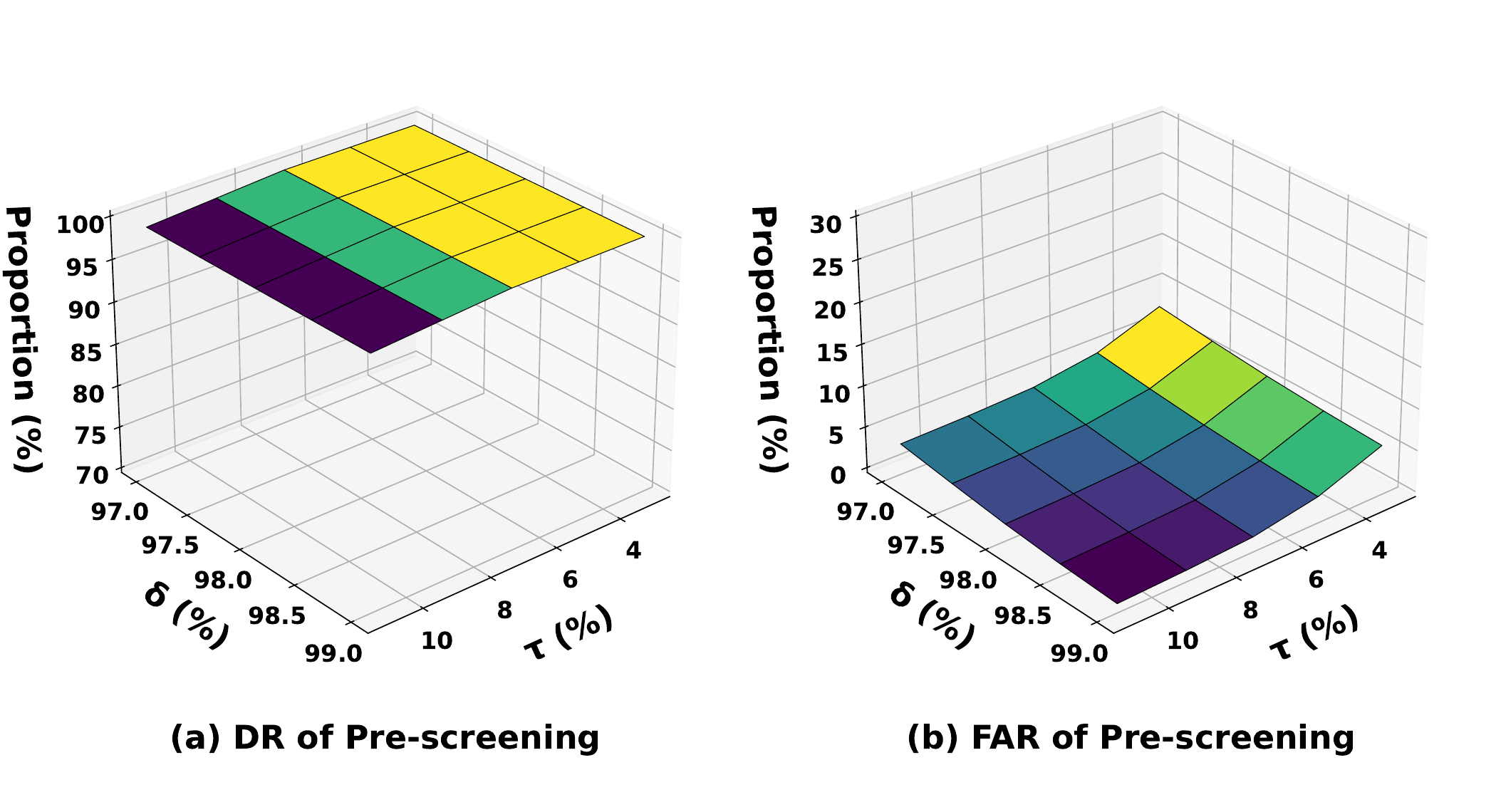}}
    \caption{\centering\textbf{Impact of $\delta$ and $\tau$ on Pre-screening Performance on ML-1M, averaged over six representative attacks.}}

    \label{fig:pre_screening_surface}
\end{figure}

\subsection{Overall Performance}

We evaluate the detection performance of different methods under six representative shilling attacks, reporting DR, FAR and their averages across all attack strategie. As shown in Table~\ref{tab:main_results}, PCA-VarSelect and GAGE exhibit unstable performance: although they sometimes achieve high DR, their FAR is often unacceptably high. DGA-MFCA also yields strong DR in some cases but suffers from moderate to high FAR. CBS performs reasonably well, with low FAR and decent average DR, but its results vary considerably across datasets and attacks. In contrast, the Llama-3-70B-Instruct model achieves consistently high DR and low FAR, demonstrating strong overall effectiveness. Furthermore, our SemanticShield model improves detection even further, reaching near-perfect DR with negligible FAR and achieving the best overall results. These findings highlight the robustness of LLM-based semantic auditing, particularly when enhanced with reinforcement fine-tuning.

To further assess generalization capability, SemanticShield is additionally evaluated against two attack strategies that were not included during training, namely GOAT~\cite{wu2021ready} and FedRecAttack~\cite{rong2022fedrecattack}. As shown in Table~\ref{tab:generalization}, SemanticShield consistently achieves high DR while keeping FAR at a negligible level,  indicating strong robustness to previously unseen attacks and addressing the limited generalization of traditional methods.

Beyond detection effectiveness, we assess recommendation quality using HR@50 and NDCG@50, and further examine consistency through RC$_{\text{HR}}$ and RC$_{\text{NDCG}}$, which measure the relative HR@50 and NDCG@50 compared with clean training, both in the attacked setting and after LLM auditing. All values are averaged over six representative attack strategies.
As shown in Table~\ref{tab:robustness}, both Llama-3-70B-Instruct and our SemanticShield maintain recommendation performance at nearly 100\% consistency, indicating that the ranking quality is almost unaffected. 
In addition, SemanticShield achieves slightly higher RC$_{\text{HR}}$ and RC$_{\text{NDCG}}$ across datasets, reflecting that its high detection accuracy leads to negligible impact on recommendation performance.

\subsection{Ablation Study}
Figure~\ref{fig:pre_screening_surface} shows the averaged pre-screening performance over six attacks on ML-1M, illustrating the effect of hyperparameters $\delta$ and $\tau$. The results indicate that both DR and FAR remain stable across a wide range of values. Importantly, the pre-screening stage can filter out nearly all fake users while sacrificing only a small fraction of genuine ones, ensuring a reliable candidate set $\mathcal{S}$ for subsequent LLM auditing.

\begin{figure}[t]
    \centering
    \adjustbox{center}{%
    \includegraphics[width=1.05\linewidth]{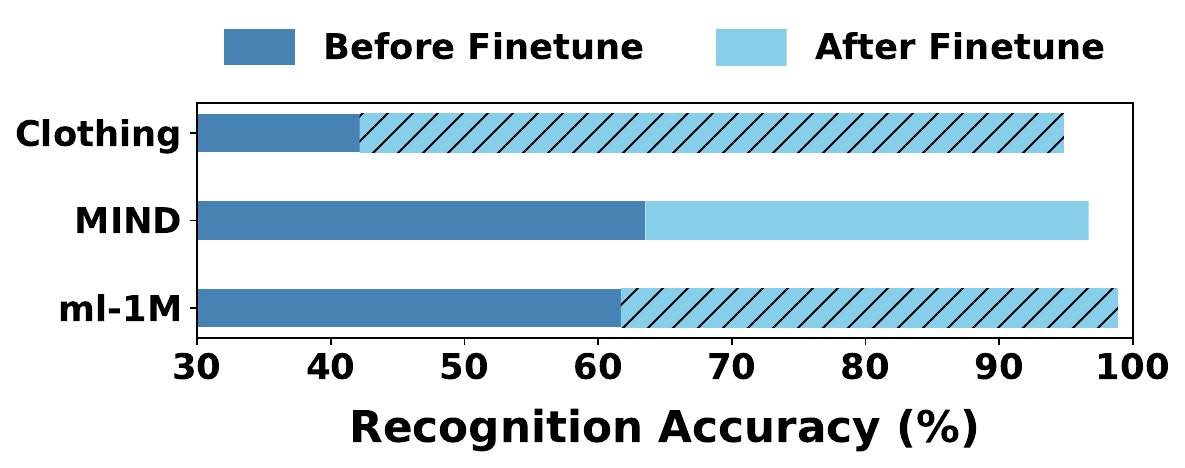}}
    \caption{\textbf{Recognition accuracy before and after finetuning.}}
    \label{fig:recognition_accuracy}
\end{figure}

To evaluate the impact of GRPO, we measure the recognition accuracy of Qwen2.5-1.5B-Instruct on the candidate set $\mathcal{S}$ before and after fine-tuning.
 As shown in Figure~\ref{fig:recognition_accuracy}, the model is tested on all three datasets, where accuracy is measured as the proportion of users in $\mathcal{S}$ whose labels are correctly identified. It is observed that fine-tuning substantially improves detection, achieving over 95\% accuracy across datasets. This finding confirms the critical role of GRPO in ensuring reliable detection.

\section{Conclusion}

In this paper, we explore the use of LLMs to detect shilling attacks in recommender systems. A two-stage framework is proposed that integrates behavioral pre-screening with LLM-based semantic auditing, addressing the limitation of prior methods that overlooked item-side semantics. Furthermore, we enhance auditing with reinforcement learning fine-tuning. Experiments on three public datasets demonstrate that our approach consistently outperforms state-of-the-art methods while maintaining strong recommendation performance, and further evaluations on unseen attack strategies confirm its strong generalization capability.

\bibliographystyle{IEEEbib}
\bibliography{strings,refs}

\begin{thebibliography}{10}

\bibitem{wu2021survey}
Le~Wu, Xiangnan He, Xiang Wang, Kun Zhang, and Meng Wang,
\newblock ``A survey on neural recommendation: From collaborative filtering to
  content and context enriched recommendation,''
\newblock {\em arXiv preprint arXiv:2104.13030}, 2021.

\bibitem{koren2021advances}
Yehuda Koren, Steffen Rendle, and Robert Bell,
\newblock ``Advances in collaborative filtering,''
\newblock {\em Recommender systems handbook}, pp. 91--142, 2021.

\bibitem{hu2008collaborative}
Yifan Hu, Yehuda Koren, et~al.,
\newblock ``Collaborative filtering for implicit feedback datasets,''
\newblock in {\em 2008 Eighth IEEE international conference on data mining}.
  Ieee, 2008, pp. 263--272.

\bibitem{he2020lightgcn}
Xiangnan He, Kuan Deng, Xiang Wang, Yan Li, et~al.,
\newblock ``Lightgcn: Simplifying and powering graph convolution network for
  recommendation,''
\newblock in {\em Proceedings of the 43rd International ACM SIGIR conference on
  research and development in Information Retrieval}, 2020, pp. 639--648.

\bibitem{gunes2014shilling}
Ihsan Gunes, Cihan Kaleli, Alper Bilge, and Huseyin Polat,
\newblock ``Shilling attacks against recommender systems: a comprehensive
  survey,''
\newblock {\em Artificial Intelligence Review}, vol. 42, no. 4, pp. 767--799,
  2014.

\bibitem{mehta2009unsupervised}
Bhaskar Mehta and Wolfgang Nejdl,
\newblock ``Unsupervised strategies for shilling detection and robust
  collaborative filtering,''
\newblock {\em User Modeling and User-Adapted Interaction}, vol. 19, no. 1, pp.
  65--97, 2009.

\bibitem{zhou2014detection}
Wei Zhou, Yun~Sing Koh, Junhao Wen, Shafiq Alam, and Gillian Dobbie,
\newblock ``Detection of abnormal profiles on group attacks in recommender
  systems,''
\newblock in {\em Proceedings of the 37th international ACM SIGIR conference on
  Research \& development in information retrieval}, 2014, pp. 955--958.

\bibitem{dou2017collaborative}
Tong Dou, Junliang Yu, Qingyu Xiong, Min Gao, Yuqi Song, and Qianqi Fang,
\newblock ``Collaborative shilling detection bridging factorization and user
  embedding,''
\newblock in {\em International conference on collaborative computing:
  networking, applications and worksharing}. Springer, 2017, pp. 459--469.

\bibitem{xu2024detecting}
Yishu Xu, Peng Zhang, Hongtao Yu, et~al.,
\newblock ``Detecting group shilling attacks in recommender systems based on
  user multi-dimensional features and collusive behaviour analysis,''
\newblock {\em The Computer Journal}, vol. 67, no. 2, pp. 604--616, 2024.

\bibitem{song2020poisonrec}
Junshuai Song et~al.,
\newblock ``Poisonrec: an adaptive data poisoning framework for attacking
  black-box recommender systems,''
\newblock in {\em 2020 IEEE 36th international conference on data engineering
  (ICDE)}. IEEE, 2020, pp. 157--168.

\bibitem{sun2019research}
Zhu Sun et~al.,
\newblock ``Research commentary on recommendations with side information: A
  survey and research directions,''
\newblock {\em Electronic Commerce Research and Applications}, vol. 37, pp.
  100879, 2019.

\bibitem{zhao2023survey}
Wayne~Xin Zhao, Kun Zhou, Junyi Li, Tianyi Tang, Xiaolei Wang, Yupeng Hou,
  Yingqian Min, et~al.,
\newblock ``A survey of large language models,''
\newblock {\em arXiv preprint arXiv:2303.18223}, 2023.

\bibitem{wu2024survey}
Likang Wu, Zhi Zheng, Zhaopeng Qiu, Hao Wang, Hongchao Gu, Tingjia Shen, Chuan
  Qin, Chen Zhu, Hengshu Zhu, Qi~Liu, et~al.,
\newblock ``A survey on large language models for recommendation,''
\newblock {\em World Wide Web}, vol. 27, no. 5, pp. 60, 2024.

\bibitem{zhao2024recommender}
Zihuai Zhao, Wenqi Fan, Jiatong Li, Yunqing Liu, Xiaowei Mei, et~al.,
\newblock ``Recommender systems in the era of large language models (llms),''
\newblock {\em IEEE Transactions on Knowledge and Data Engineering}, vol. 36,
  no. 11, pp. 6889--6907, 2024.

\bibitem{abdi2010principal}
Herv{\'e} Abdi and Lynne~J Williams,
\newblock ``Principal component analysis,''
\newblock {\em Wiley interdisciplinary reviews: computational statistics}, vol.
  2, no. 4, pp. 433--459, 2010.

\bibitem{chu2025sft}
Tianzhe Chu et~al.,
\newblock ``Sft memorizes, rl generalizes: A comparative study of foundation
  model post-training,''
\newblock {\em arXiv preprint arXiv:2501.17161}, 2025.

\bibitem{shao2024deepseekmath}
Zhihong Shao, Peiyi Wang, et~al.,
\newblock ``Deepseekmath: Pushing the limits of mathematical reasoning in open
  language models,''
\newblock {\em arXiv preprint arXiv:2402.03300}, 2024.

\bibitem{qwen2.5}
Qwen Team,
\newblock ``Qwen2.5: A party of foundation models,'' September 2024.

\bibitem{llama3modelcard}
AI@Meta,
\newblock ``Llama 3 model card,''
\newblock 2024.

\bibitem{rendle2012bpr}
Steffen Rendle, Christoph Freudenthaler, Zeno Gantner, et~al.,
\newblock ``Bpr: Bayesian personalized ranking from implicit feedback,''
\newblock {\em arXiv preprint arXiv:1205.2618}, 2012.

\bibitem{harper2015movielens}
F~Maxwell Harper and Joseph~A Konstan,
\newblock ``The movielens datasets: History and context,''
\newblock {\em Acm transactions on interactive intelligent systems (tiis)},
  vol. 5, no. 4, pp. 1--19, 2015.

\bibitem{wu2020mind}
Fangzhao Wu et~al.,
\newblock ``Mind: A large-scale dataset for news recommendation,''
\newblock in {\em Proceedings of the 58th annual meeting of the association for
  computational linguistics}, 2020, pp. 3597--3606.

\bibitem{hou2024bridging}
Yupeng Hou, Jiacheng Li, Zhankui He, An~Yan, et~al.,
\newblock ``Bridging language and items for retrieval and recommendation,''
\newblock {\em arXiv preprint arXiv:2403.03952}, 2024.

\bibitem{lam2004shilling}
Shyong~K Lam and John Riedl,
\newblock ``Shilling recommender systems for fun and profit,''
\newblock in {\em Proceedings of the 13th international conference on World
  Wide Web}, 2004, pp. 393--402.

\bibitem{zhang2022pipattack}
Shijie Zhang, Hongzhi Yin, Tong Chen, Zi~Huang, Quoc Viet~Hung Nguyen, and
  Lizhen Cui,
\newblock ``Pipattack: Poisoning federated recommender systems for manipulating
  item promotion,''
\newblock in {\em Proceedings of the Fifteenth ACM International Conference on
  Web Search and Data Mining}, 2022, pp. 1415--1423.

\bibitem{wang2023revisiting}
Zhiye Wang, Baisong Liu, Chennan Lin, Xueyuan Zhang, Ce~Hu, Jiangcheng Qin, and
  Linze Luo,
\newblock ``Revisiting data poisoning attacks on deep learning based
  recommender systems,''
\newblock in {\em 2023 IEEE Symposium on Computers and Communications (ISCC)}.
  IEEE, 2023, pp. 1261--1267.

\bibitem{wang2024unveiling}
Zongwei Wang, Junliang Yu, Min Gao, Hongzhi Yin, Bin Cui, and Shazia Sadiq,
\newblock ``Unveiling vulnerabilities of contrastive recommender systems to
  poisoning attacks,''
\newblock in {\em Proceedings of the 30th ACM SIGKDD conference on knowledge
  discovery and data mining}, 2024, pp. 3311--3322.

\bibitem{zhang2015catch}
Yongfeng Zhang, Yunzhi Tan, Min Zhang, Yiqun Liu, Tat-Seng Chua, and Shaoping
  Ma,
\newblock ``Catch the black sheep: Unified framework for shilling attack
  detection based on fraudulent action propagation.,''
\newblock in {\em IJCAI}, 2015, vol.~15, pp. 2408--2414.

\bibitem{zhang2020graph}
Fuzhi Zhang et~al.,
\newblock ``Graph embedding-based approach for detecting group shilling attacks
  in collaborative recommender systems,''
\newblock {\em Knowledge-Based Systems}, vol. 199, pp. 105984, 2020.

\bibitem{wu2021ready}
Fan Wu, Min Gao, Junliang Yu, Zongwei Wang, Kecheng Liu, and Xu~Wang,
\newblock ``Ready for emerging threats to recommender systems? a graph
  convolution-based generative shilling attack,''
\newblock {\em Information Sciences}, vol. 578, pp. 683--701, 2021.

\bibitem{rong2022fedrecattack}
Dazhong Rong et~al.,
\newblock ``Fedrecattack: Model poisoning attack to federated recommendation,''
\newblock in {\em 2022 IEEE 38th International Conference on Data Engineering
  (ICDE)}. IEEE, 2022, pp. 2643--2655.

\end{thebibliography}

\end{document}